\documentclass[conference]{IEEEtran}
\IEEEoverridecommandlockouts
\usepackage{cite}
\usepackage{CJKutf8}
\usepackage{amsmath,amssymb,amsfonts}
\usepackage{algorithmic}
\usepackage{graphicx}
\usepackage{textcomp}
\usepackage{multirow}
\usepackage{xcolor}
\def\BibTeX{{\rm B\kern-.05em{\sc i\kern-.025em b}\kern-.08em
    T\kern-.1667em\lower.7ex\hbox{E}\kern-.125emX}}
\begin{document}

\title{Bridging the Language Gap: Knowledge Injected Multilingual Question Answering\\
}

\author{
\IEEEauthorblockN{1\textsuperscript{st} Zhichao Duan}
\IEEEauthorblockA{\textit{Department of Computer Science \& Technology} \\
\textit{Tsinghua University}\\
Beijing, China \\
dzc20@mails.tsinghua.edu.cn}
\\
\IEEEauthorblockN{3\textsuperscript{rd} Zhengyan Zhang}
\IEEEauthorblockA{\textit{Department of Computer Science and Technology} \\
\textit{Tsinghua University}\\
Beijing, China \\
zy-z19@mails.tsinghua.edu.cn}
\\
\IEEEauthorblockN{5\textsuperscript{th} Ning Liu}
\IEEEauthorblockA{\textit{School of Software} \\
\textit{Shandong University}\\
Jinan, China \\
victorliucs@gmail.com}
\and
\IEEEauthorblockN{2\textsuperscript{nd} Xiuxing Li}
\IEEEauthorblockA{\textit{Department of Computer Science \& Technology} \\
\textit{Tsinghua University}\\
Beijing, China \\
lxx16@mails.tsinghua.edu.cn}
\\
\IEEEauthorblockN{4\textsuperscript{th} Zhenyu Li}
\IEEEauthorblockA{\textit{Department of Computer Science and Technology} \\
\textit{Tsinghua University}\\
Beijing, China \\
zy-li21@mails.tsinghua.edu.cn}
\\
\IEEEauthorblockN{6\textsuperscript{th} Jianyong Wang}
\IEEEauthorblockA{\textit{Department of Computer Science and Technology} \\
\textit{Tsinghua University}\\
Beijing, China \\
jianyong@mail.tsinghua.edu.cn}
}

\maketitle

\begin{abstract}
% Pretrained language models have exhibited remarkable performances in various tasks in natural language area, which is essential and fundamental for many downstream applications. However, the performance of multilingual pretrained language model is highly limited by the amount of multilingual data. Existing methods usually use sentences which include entities to incorporate knowledge into pretrained language models. These kinds of methods can't be directly applied in a multilingual scenario since there isn't many aligned texts. 
% In this work, we use multilingual knowledge triples to inject knowledge into pretrained language models by doing entity completion explicitly. 
% In this way, the model performance can be improved without large aligned texts. Results on MLQA, a multilingual question answering dataset, demonstrate that proposed method can improve the performance by a large margin.
% monolingual question answering due to its complexity, especially in the cross-lingual setting. This task is to find the answer to a specific problem in a context. 
Question Answering (QA) is the task of automatically answering questions posed by humans in natural languages. There are different settings to answer a question, such as abstractive, extractive, boolean, and multiple-choice QA. As a popular topic in natural language processing tasks, extractive question answering task (extractive QA) has gained extensive attention in the past few years. With the continuous evolvement of the world, generalized cross-lingual transfer (G-XLT), where question and answer context are in different languages, poses some unique challenges over cross-lingual transfer (XLT), where question and answer context are in the same language. With the boost of corresponding development of related benchmarks, many works have been done to improve the performance of various language QA tasks. However, only a few works are dedicated to the G-XLT task. In this work, we propose a generalized cross-lingual transfer framework to enhance the model's ability to understand different languages. Specifically, we first assemble triples from different languages to form multilingual knowledge.  Since the lack of knowledge between different languages greatly limits models' reasoning ability, we further design a knowledge injection strategy via leveraging link prediction techniques to enrich the model storage of multilingual knowledge. In this way, we can profoundly exploit rich semantic knowledge. Experiment results on real-world datasets MLQA demonstrate that the proposed method can improve the performance by a large margin, outperforming the baseline method by 13.18\%/12.00\% F1/EM on average. Our code and models will be open-sourced.

% indicates that our method can ourperform the baseline by over 10\% on average in the tested cases
% demonstrate that the proposed method can improve the performance by 13.18\% on F1 score and 12.00\% on EM score on average in the tested cases.

% It is vital to many real-world applications. However, the performance of many state-of-the-art methods is far behind satisfaction under the cross-lingual setting. Existing methods try to put different entities back to sentences and use PLMs to restore them. These methods dumped entities that do not have semantic-related sentences. 
% In this work, we use multilingual knowledge triples to bridge the gaps between different forms of representations and improve the performance of PLMs in cross-lingual settings. In this way, the off-the-shelf knowledge triples can be directly leveraged to improve the model performance. Results on MLQA, a multilingual question answering dataset, demonstrate that the proposed method can improve the performance by a large margin.
\end{abstract}


\section{Introduction}
% Question answering task have gained more and more attention during the past few years. It has many real world applications like expert systems and online QA software. This task is to find the answer span in a context according to the problem. Many works have been done to improved the performance in the monolingual settings, i.e., the problem and the context are in the same language representation. However, the multilingual question answering, especially cross-lingual question answering remains a challenging task. 

% In multilingual question answering, the context and the problem can be in the same or different languages. When the context and the problem are in different language representations, the ability to aligning different languages is required to conclude the final answer. With the advanced development of pretrained language models, many recent works use finetuned PLMs to solve multilingual question answering problem. PLMs use unsupervised objective to learn the contextualized representation of words and are proven to be sucessful in many downstream tasks like text classification and named entity recognition. 

Extractive question answering task (extractive QA) is to find the answer span in the given context according to the problem, which is potentially a challenging task that can drive the development of methods for Natural Language Processing (NLP). Modern QA algorithms use Pre-trained Language Models (PLMs) \cite{b1,b0,b2,b3,b4} to infer the answer span. 
PLMs can capture different levels of features and generate informative representations by leveraging unsupervised training objectives and enormous training text. They have demonstrated powerful performance on many NLP tasks like sentiment classification \cite{b4} and relation extraction \cite{b5}. 

The essence of QA task is to represent the question and context precisely~\cite{bb7}. Hence, how to improve the representation ability of the model has become a critical step to address the problem. Recent works mainly focus on English QA problems, such as SQuAD~\cite{b6,b7} and cross-lingual QA task (XLT)~\cite{b8}. Some other works only view QA task as an indicator to evaluate the model performance~\cite{bluke,bt5,bdice,brein}. Under these circumstances, data can be plenty for high-resource languages like English, Chinese, German, Spanish, and French. Still, both collection and proliferation of data are insufficient for low-resource languages like Urdu. Even when a large language model is pre-trained on large amounts of multilingual data, languages like English can contain orders of magnitude more data in common sources for pretraining like Wikipedia. 
\begin{figure}[h]
\centerline{\includegraphics[width=10cm]{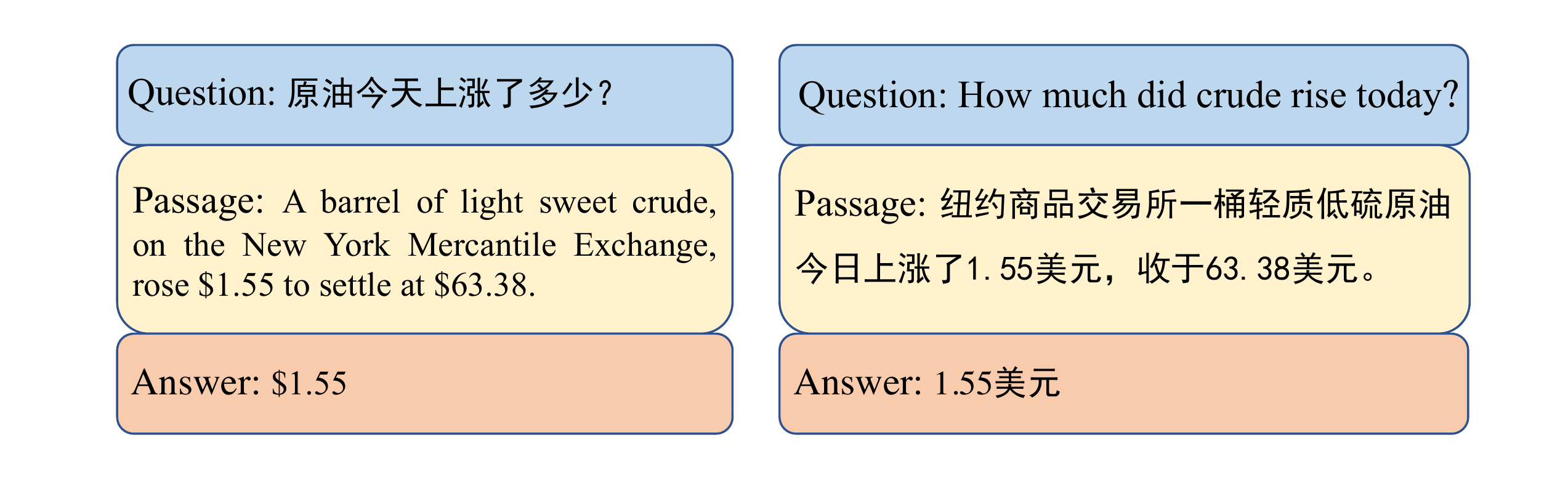}}
\caption{A question-answering sample where context is in English while question is in Chinese and the reverse. Answer text is extracted from the passage according to the question.
}
\label{sample}
\end{figure}
Therefore, the mainstream approaches to tackle the English QA problem is to improve the representation ability of the model through introducing high-resource languages.

However, the generalized cross-lingual QA task (G-XLT), where problem and context are in different language representations, remains an open problem due to its complexity. Compared to XLT, G-XLT poses some unique challenges. Specifically, the latter task requires the QA model to align different representations automatically, understand the question, reason over the context, and conclude the final answer. This problem can not be solved simply by a large PLM with knowledge in each language individually but without understanding different languages and their connections. The QA model needs to be aware of different representations of the same word. For example, Fig.~\ref{sample} gives a demonstration of cross-lingual QA existing in MLQA benchmark \cite{b9}, as humans, we can only find the answer when we understand the question, and so do PLMs. Surely we can first translate the question into English and then find the answer span~\cite{b9}, but this method normally uses another PLM to handle the translation process and causes error accumulation. 

There are some works aiming to enhance the awareness of different entities in PLMs to boost performance in monolingual NLP tasks, which are usually English~\cite{b10,b11,b12,b13,b14}. Reference~\cite{b15} uses extra layers called adapters to learn the knowledge introduced by each pretraining task and improves the performance of PLMs on multiple downstream tasks. Methods like~\cite{b16,b17} explore different ways to take advantage of texts that contain rich knowledge. Reference~\cite{b16} firstly transfers knowledge triples into text and then utilizes knowledge-enriched text to finetune PLMs further. Replacing detection and entity restoring are studied in~\cite{b17}. Rather than injecting new knowledge, some works try to regularize the embedding space and thus benefit downstream tasks. Reference~\cite{b14} proposes a method that adopts a siamese network to make the output embedding space more consistent. Then they build an analogy dataset and use extra training objectives to ensure that word representations are globally consistent. Instead of using analogy training, ~\cite{b18} uses Wikipedia entities to ground representations of different forms of entities (same entity in different languages).  In general, they either design entity-related auxiliary objectives or use corpus that contains rich semantic knowledge to finetune PLMs. These methods can be applied to inject mono-lingual knowledge into PLMs. However, without the alignment information between different languages, it is hard for the model to realize the various representations of a single word or sentence and better solve the G-XLT task. Unfortunately, it is also quite expensive to build multilingual aligned corpus. Besides, they only use entities that have semantically related sentences and dumped those entities without related sentences. These methods highly limited the application of aligned resources. When it comes to cross-lingual QA task, we can't simply use any of these methods to enhance the performance due to insufficient data and applicable conditions.

To solve the aforementioned problems, we propose a novel G-XLT framework. Firstly, we mix triples from different languages to represent multilingual knowledge. In this step, we also assemble monolingual triples to avoid catastrophic forgetting. In the next, we adopt link prediction techniques to inject multilingual knowledge into the PLMs. In this way, we can enrich the knowledge stored in each language and smooth the knowledge transfer between different languages. Meanwhile, the reasoning ability of our model can be improved through link prediction. Finally, our framework is tested on MLQA~\cite{b9}, a cross-lingual extractive QA benchmark. Results show that our method can outperform the baseline method by over 10\% improvement on average in tested cases. 

Our contributions are listed as follows:
\begin{itemize}
\item We propose a generalized cross-lingual transfer framework which can better understand different languages by exploiting multilingual knowledge bases.
\item We design a knowledge injection technique using link prediction to enrich the model storage of multilingual knowledge and therefore enhance performance in G-XLT task.
\item We demonstrate our results and prove our method can not only promote the knowledge transfer between different languages but also enhance the reasoning ability in G-XLT task.
\end{itemize}

\begin{figure*}[h]
\centerline{\includegraphics[width=18cm]{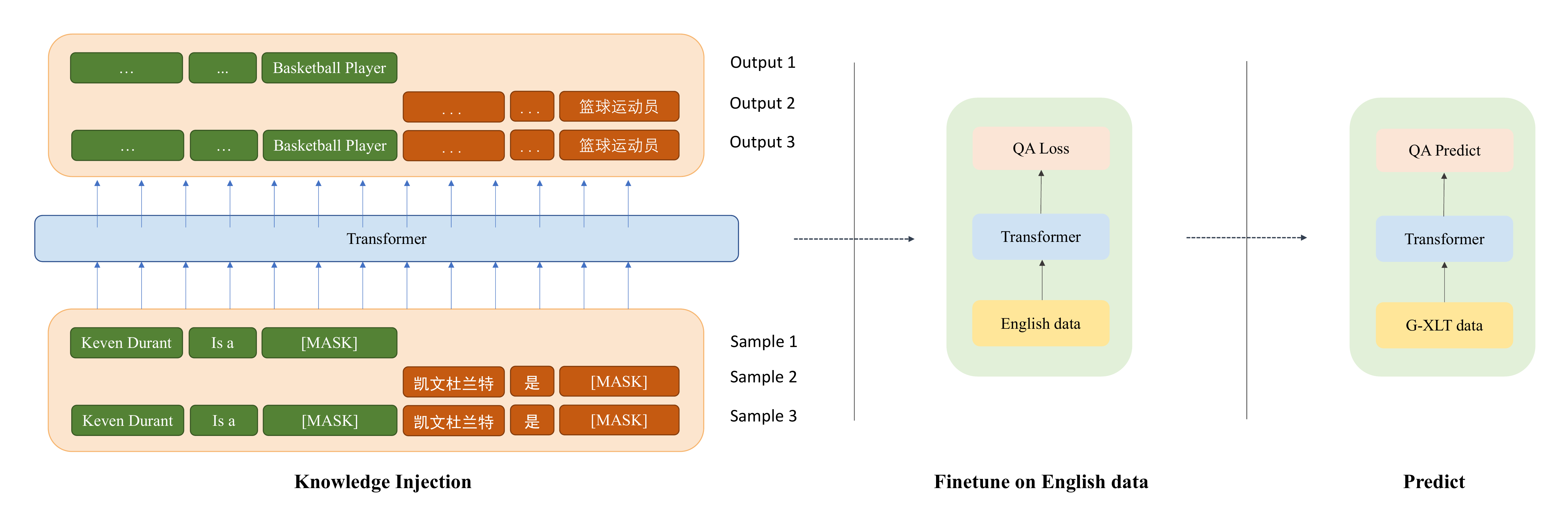}}
\caption{The overall structure of our method. We demonstrate the case where the tail entities are masked in all three samples. Note that when we actually assemble training triples, we separately mask out head entity and tail entity. Sample 1 is in English and sample 2 is in Chinese and sample 3 is mixed with English and Chinese. 3 samples are feed as input to the PLMs and the target predictions are two representations of the entity "Basketball Player". Note that we only limit the output of [MASK]] tokens. After the knowledge injection step, the PLM is finetuned on the English data in MLQA and finally will be used to make predictions based on G-XLT data.  }
\label{overall}
\end{figure*}

\section{Related Work}
% Note that G-XLT is based on the special setting of the QA task. As mentioned earlier, a potent way to solve the QA problem is to use finetuned PLMs. Even with the great success of PLMs in many NLP tasks, previous works implied that we could enhance different abilities of PLMs from different angles. Therefore, we can improve the performance on G-XLT task by enhancing abilities of PLMs. Since we already briefly talked about enhanced PLMs in the previous section, we will skip this part in this section and only talk about PLMs based link prediction (enhancing mechanism used in our work) and multilingual question answering.

% The goal of our work is to better solve G-XLT using PLMs which will be introduced in the following parts.
With the wide popularization of PLMs, many modern QA algorithms use PLMs to encode question and context and conduct classification based on extra layers. In our work, we follow this procedure. Therefore, we will introduce two parts in this section which are multilingual question answering task along with its two categories and pretrained language models.

\subsection{Multilingual question answering}
QA is a hot topic in NLP area. There are massive datasets available, and these datasets pose different challenges to QA algorithms \cite{b6, b23, b24}. One of these challenges is to find the answer span when the context and the problem are in different languages, and this task is called generalized cross-lingual transfer(G-XLT). More formally, cross-lingual transfer problem requires a model to identify answer $a_x$ in context $c_x$ according to problem $q_x$ where $x$ is the language used. Meanwhile, generalized cross-lingual transfer requires a model to find the answer span $a_z$ in context $c_z$ according to question $q_y$ where $z$ and $y$ are languages used. Reference \cite{b9} uses a designed pipeline to construct a large parallel multilingual QA dataset that contains 7 languages. With the rapid development of PLMs and remarkable success in various fields of natural language processing, adopting PLMs to encode input text and using extra layers to conduct classification has gradually become a popular approach to solve multilingual QA task. Extensive experiments using state-of-the-art PLMs are reported in \cite{b9}. Their results show that there are quite large room for improvement on both XLT and G-XLT tasks. One of the major challenges in G-XLT task is to enhance the connection between different languages. Reference \cite{b18} uses hyperlink prediction task to firstly enhance the connection between different representations and then finetune the model on the target task data in English. Finally, they evaluate the performance on all languages. It is worth noting that this method creatively uses hyperlink as the bridge of different languages.

% \subsection{Enhanced PLMs}
% Reference~\cite{b14} proposes a new framework by using analogy training to make the word embedding spaces more consistent. They adopt a siamese network to better model the relation between analogies to form global consistency in word embeddings. Extensive experiments show that their method has effective performance in tasks that require global consistency of word embeddings. 
% Reference~\cite{b16} firstly transforms knowledge triples into sentences and uses transformed sentences to further finetune the PLM. This method stores knowledge into sentences and adopts masked language modeling objective~\cite{b1} to learn it. Reference~\cite{b19} is based on two tasks: entity discrimination task and entity discrimination task, which can help PLMs to understand entities and relations in text deeply. In our work, we use the link prediction task as an intermediate step to inject multilingual knowledge into PLMs and enhance the understanding of different languages. 

\subsection{Pretrained Language Models}
PLMs such as BERT~\cite{b1}, have gained great success in various downstream tasks. They apply unsupervised training objective to make full use of the huge text resources. Typically, they use stacked transformers as model structure and randomly mask out some tokens in each sample. The goal of PLMs is to restore the original text. Later on, some new mask strategies and more complex models were proposed to improve the performance on many tasks. PLMs contain rich world knowledge, which could be used as an important complement to highly limited knowledge bases. Some works have been done to use PLMs to do link prediction task~\cite{b20,b21}. Knowledge Graph (KG) consists of entities $E$, relations $R$ and triplets $(h,r,t)$ where $h$ and $t$ are in entity set $E$ and $r$ is in relation set $R$. Link prediction is to find the appropriate entity that corresponds to another entity linked by a specific relation. That is to say, complete $(h,r,?)$ or $(?,r,t)$.

% It can be quite useful especially in the inductive setting, i.e., the testing triples may not appear in the training triples. 
PLMs contain meaningful representations of different entities and relations. Reference \cite{b21} explores leveraging knowledge inside PLMs on link prediction tasks. Reference \cite{b22} uses PLMs to generate entity and relation embeddings and continuously improves them using MLM objective. They train BERT as a classifier to examine whether a specific triple is correct or not. That is to say, BERT will check each possible $(?, r, t)$ and $(h, r, ?)$ and determine the final matching entity. This method costs tremendous time and resources due to its traversal operation. Later on, \cite{b20} proposes to use several [MASK] tokens to represent the missing entity and rank all possible candidates at once to reduce the time consumption in both training and testing. Their work proves that PLMs have the ability to infer missing entities and we adopt this method to inject knowledge into PLMs.

\section{Proposed Method}
We propose a new G-XLT framework through enhancing the reasoning abilities and the understanding of different languages of PLMs. The entire method consists of three parts: (1) cross-lingual triple assembling; (2) multilingual knowledge injection; (3) final finetuning on English data. The overall architecture is shown in Fig.~\ref{overall}.

\subsection{Cross-lingual triple assembling}
The information of our world can be represented in the form of knowledge triples. As mentioned before, we can use $(h,r,t)$ to represent a specific triple where $h$ and $t$ belong to entity set $E$ and $r$ belongs to relation set $R$. Here, we extend this concept into multilingual cases. We use $E$ and $R$ to represent entity sets and relation sets separately, which include entities and relations in different languages. We use $(h_i,r_i,t_i)$ to represent a triple in language $i$. We sampled 20,000 triples in Wikidata. 7 languages are considered in this work due to our benchmark MLQA only has data in 7 languages. Three types of triples are built in this phase. The first kind of triples are $(h_i, r_i, t_i)$ like ones where head entity, tail entity and relation are from the same language. For example, as shown in Fig.~\ref{overall}, ``(Keven Durant, is a, Basketball Player)'' and ``\begin{CJK}{UTF8}{gbsn}(凯文杜兰特，是，篮球运动员)\end{CJK}'' are two types of representations of the same triple and are assembled in this stage. $(h_j, r_i, t_i)$ and $(h_i, r_i, t_j)$ are the second kind of triples where head entity or tail entity are in another language. We can combine the previous samples and form mixed data like ``\begin{CJK}{UTF8}{gbsn}(凯文杜兰特，是，Basketball Player).\end{CJK}'' $(h_i, r_i, t_i, h_j, r_j, t_j)$ is the third kind of triples where two representations are concatenated to form one line of data like ``\begin{CJK}{UTF8}{gbsn}(Keven Durant, is a, Basketball Player, 凯文杜兰特，是，篮球运动员)\end{CJK}''. These three kinds of triples are randomly assembled and used to perform link prediction in the next step.

\begin{figure*}[h]
\centerline{\includegraphics[width=15cm]{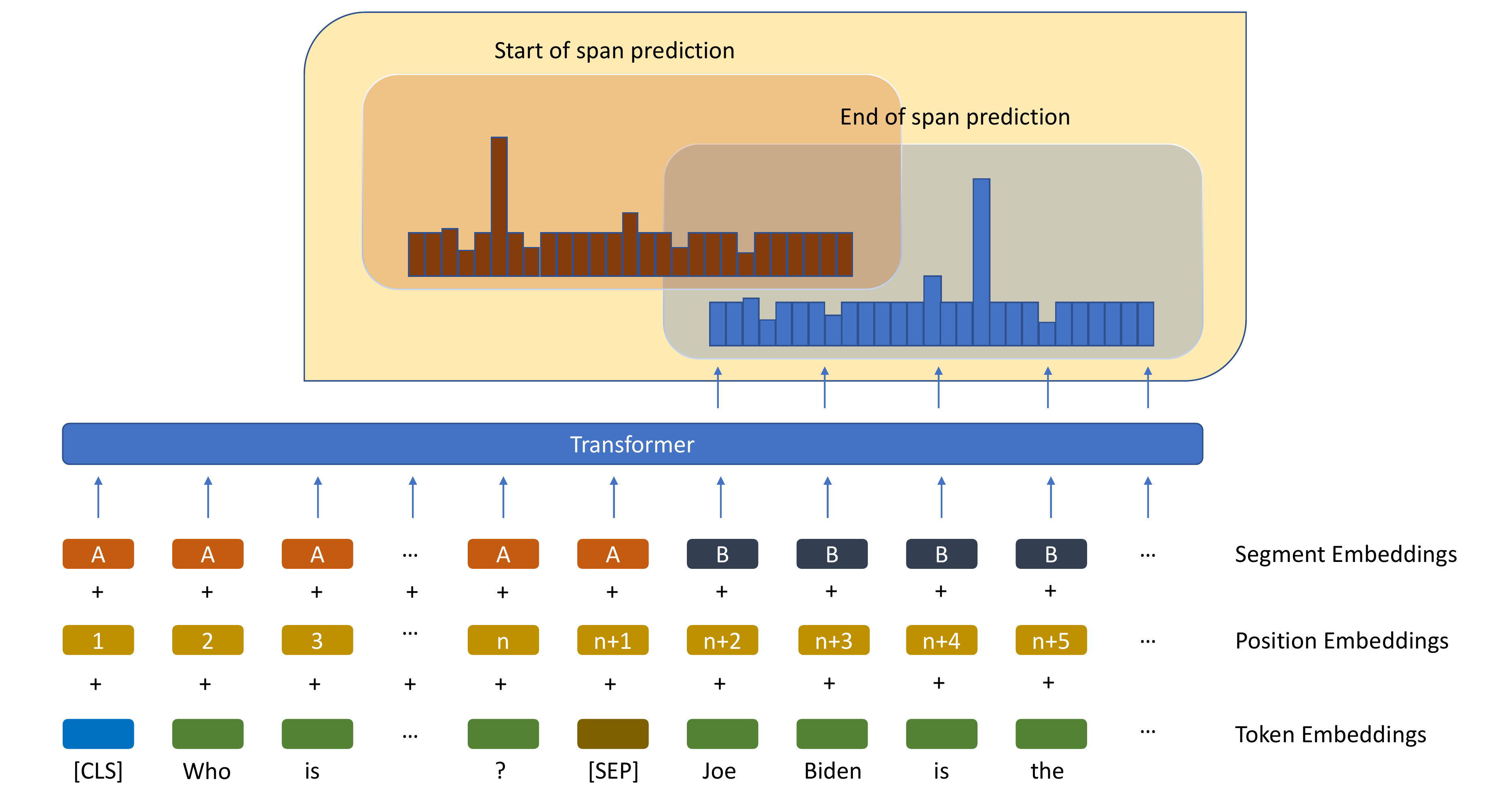}}
\caption{Illustration of how we finetuning PLMs on QA data. Here we put question in the left side and put context in the right. Segment embeddings  are used to distinguish between question and context.}
\label{finetune}
\end{figure*}

\subsection{Knowledge Injection}
After generating the knowledge triples, we design a knowledge injection strategy via link prediction techniques to inject multilingual knowledge and align different representations. As we mentioned before, one of many reasons PLMs achieved great success is that they use unsupervised training objective to optimize the entire model. To be more specific, they use masked language modeling method to build PLMs based on large unlabeled text. Masked language modeling usually predicts the missing tokens based on the given ones by maximizing the log-likelihood:

\begin{equation}\label{eq-mlm}
    \mathcal{L}_{mlm}(\theta) = \sum\limits_{i=1}^{n}m_ilogp(x_i|\boldsymbol{x}{\circ}\boldsymbol{m};\boldsymbol{\theta})
\end{equation}

$\boldsymbol{x}$ is the sequence of tokens derived from a sentence, $\circ$ is the operation of element-wise production, $\boldsymbol{m}$ is an indicator that shows which tokens in the sequence are replaced with [MASK] tokens and $\theta$ is the parameters of the model.

Here, a key part of our method is that we use PLMs to predict the missing entities when we mask out the head entities or the tail entities of triples. That is to say,
the goal of our model is to find the matching entity given relation and another entity. Take the example in Fig.\ref{overall}, the [MASK] token represents the missing entity ``Basketball Player'' and its Chinese version ``\begin{CJK}{UTF8}{gbsn}篮球运动员\end{CJK}'' and the goal of our model is to restore those entities.

Note that masking strategy varies according to the triple type. We mask head entity and tail entity respectively in the first kind of triples. For the first kind of triples $(h_i, r_i, t_i)$, two masked triples $(h_i, r_i, ?)$ and $(?, r_i, t_i)$ are formed and the goal of our model is to restore $t_i$ and $h_i$ respectively. Only one language appears in each sample of the first kind, which mimics the training procedure of the original masked language modeling. In this way, we can not only avoid the catastrophic forget \cite{b222}, but also enhance the reasoning ability through link prediction.
For triple $(h_j, r_i, t_i)$ and triple $(h_i, r_i, t_j)$ where the head entity and the tail entity are represented in language $j$ respectively and the rest is represented in language $i$, we form two masked triple $(?, r_i, t_i)$ and $(h_i, r_i, ?)$ accordingly. The goal of our model is to restore $h_j$ and $t_j$. 
For a triple $(h_i, r_i, t_i, h_j, r_j, t_j)$ where two forms of the same triple are concatenated to form one triple, we form two masked triples $(?, r_i, t_i, ?, r_j, t_j)$ and $(h_i, r_i, ?, h_j, r_j, ?)$. The goal of this step and the previous one is to align different representations of the same entity and make the representation more consistent in the feature space. Furthermore, we mix different languages in each sample of the second and the third kind of triples. By explicitly regularizing the representations, the model can better leverage the rich knowledge stored in high-resource language and internalize it into model parameters. 

The entity completion loss is defined as:

\begin{equation}\label{eq-mlm}
    \mathcal{L}_{ec}(\theta) = \sum\limits_{i=1}^{n}logp(e|e^{\prime},r;\boldsymbol{\theta})
\end{equation}

where $e$ is the missing entity, $e^{\prime}$ is the given entity and r is the relation, and $\theta$ is the model parameters.

\subsection{Finetune on MLQA English data}
After the intermediate step above, we will finetune PLM to make it more suitable for QA task because the different form of QA task and MLM task. This step is the most adopted approach to finetune PLM on QA task \cite{b1} and is shown in Fig.~\ref{finetune}.

Basically, we concatenate question and the context together and add segmentation embedding to two parts respectively. 
A special [SEP] token is added between question and the context to help the model distinguish the two parts. Then, tokenized samples are feedforward through PLM, and the output in each position is a vector. We use extra layers on top of PLM to determine the start span and end span of the final answer.

% \usepackage[normalem]{ulem}
\begin{figure*}[h]
\centerline{\includegraphics[width=18.2cm]{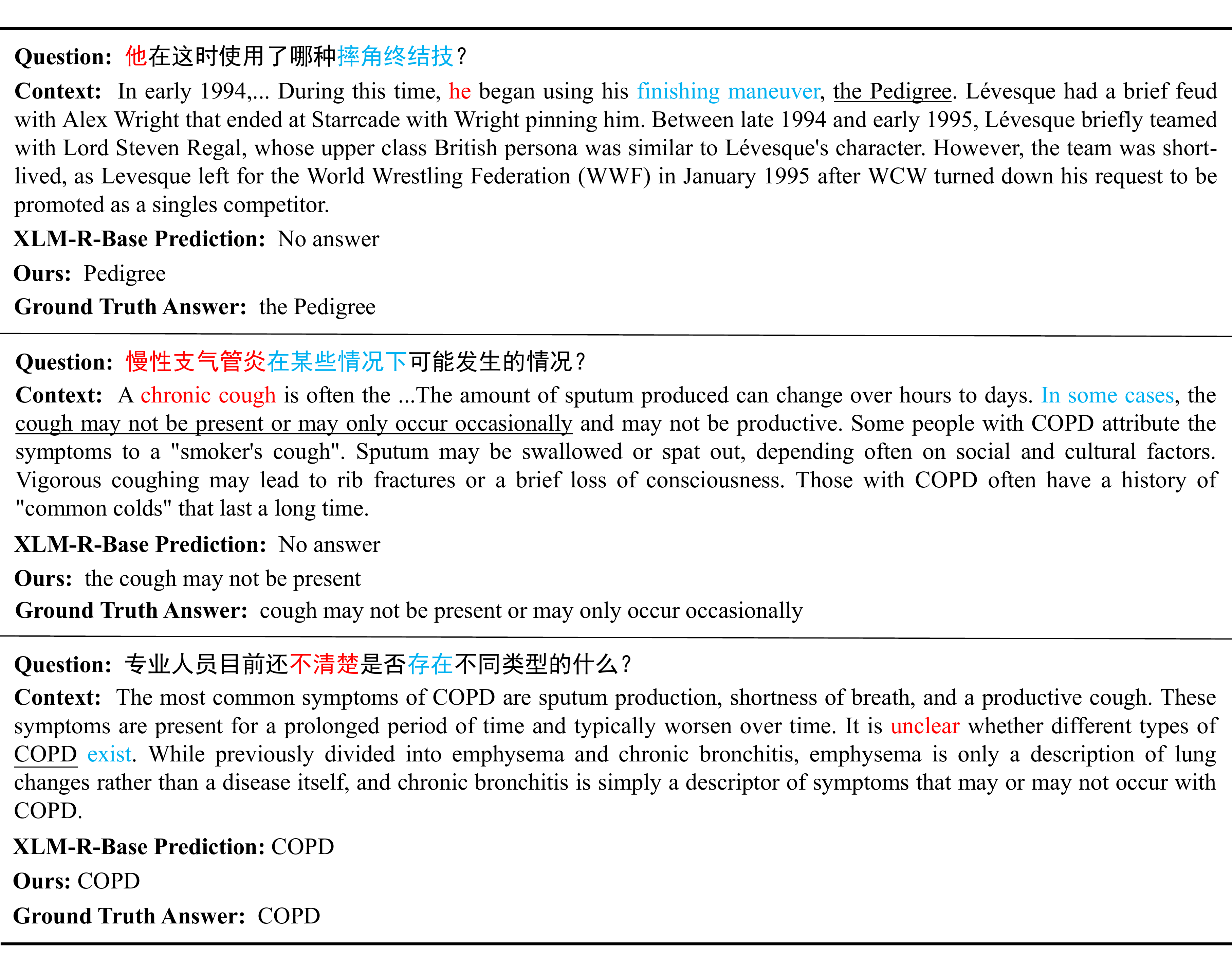}}
\caption{A case study for our method. The goal of our model is to find the answer text according to the problem and the question. \textbf{XLM-R-Base} is the baseline method we use for comparison. Characters marked in the same color have the same meaning and underlined words are the answer text.}
\label{study}
\end{figure*}

\section{Experiments}
\subsection{Experiment Setup}
\textbf{Dataset}
We evaluate our model's performance on MLQA \cite{b9}, a highly-parallel multilingual extractive QA dataset. MLQA has QA samples in 7 languages including English, Arabic, German, Spanish, Hindi, Vietnamese and Simplified Chinese. This dataset has 12,738 extractive QA instances in English and over 5K in target language. 9,019 instances are 4-way parallel, 2,930 are 3-way parallel and 789 are 2-way parallel.  MLQA is constructed from Wikipedia articles which have similar meaning in parallel languages. Labeling work is done by crowd-sourcing and translation is done by professional translators. In total, there are over 46,000 QA instances. MLQA provides data for G-XLT experiments and the dev/test split is already given.

\textbf{Baseline Methods}
Our method is built on top of XLM-R. It is proposed in~\cite{b3} and is based on the idea that pre-training multilingual language models at scale lead to significant performance improvement for a wide range of cross-lingual transfer tasks. XLM-R is trained on 2 terabytes of crawled data and covers over 100 languages. Compared to other multilingual models, it does not require language type tensors in the input. Using a relatively large vocabulary, XLM-R can detect the correct language from the input ids. Since the goal of our experiments is to demonstrate the effectiveness, we use a smaller version of XLM-R, XLM-R-Base.

\textbf{Evaluation Metrics}
The answer-preprocessing operations adopted in SQuAD~\cite{b6,b7} are targeting English QA evaluation. We use methods proposed in~\cite{b9} to preprocess data for a fair comparison. We use Exact Match (EM) and mean token F1 score as performance metrics like other literature.

\textbf{Implementation Details}
There are some multilingual PLMs like \cite{b3,b2,b0001}. For fast experiment, XLM-R-Base \cite{b3} is used as the encoder. Our method is implemented using Pytorch and Huggingface's Transformers library\cite{b25}.
which is optimized by AdamW \cite{b26}. In the knowledge injection phase, linear warmup strategy \cite{b27} is adopted for the first 6$\%$ steps and learning rate is set to 2e-5. Batch size is set to 24 and only 1 training epoch. After the knowledge injection step, we still use AdamW to finetune XLM-R-Base except that the learning rate is set to 3e-5 and the batch size is set to 16 with 2 training epochs. All the above training is done in a single RTX 3090 GPU.

\subsection{Main Results}
Since English data constituted the largest corpus among all the training data, we assume XLM-R-Base understands English better than other languages. In the QA task, we concatenate question and context together as input and feed it to the PLM. 
PLMs can accurately encode text to a certain extent. As the length of text increases, the precision of encoding will drop.
In order to see the effects our method has, we only compare performance when the context is English since context is usually much longer than the question, and results in~\cite{b3} prove that XLM-R understands English better than any other language. 

% \usepackage{multirow}
% \caption{Experimental results where context is set to English and question is set to six other languages. c/q column means the languages used in context/question. xlm-roberta-base is the baseline method and xlm-roberta-base enhanced is the improved version using our method.}
\begin{table}[]
\caption{Experimental results on English-context cases.}
\centering
\setlength{\tabcolsep}{4mm}{
\begin{tabular}{cccc}
\hline
Settings(c/q)          & Method                    & F1    & Exact Match \\ \hline
\multirow{2}{*}{EN/DE} & xlm-roberta-base          & 60.30 & 47.73       \\
                       & ours & \textbf{63.10} & \textbf{50.34}       \\ \hline
\multirow{2}{*}{EN/ES} & xlm-roberta-base          & 57.01 & 43.02       \\
                       & ours & \textbf{57.98} & \textbf{44.32}       \\ \hline
\multirow{2}{*}{EN/AR} & xlm-roberta-base          & 30.71 & 19.78       \\
                       & ours & \textbf{49.19} & \textbf{36.10}       \\ \hline
\multirow{2}{*}{EN/HI} & xlm-roberta-base          & 40.22 & 28.73       \\
                       & ours & \textbf{58.16} & \textbf{45.91}       \\ \hline
\multirow{2}{*}{EN/VI} & xlm-roberta-base          & 41.90 & 30.50       \\
                       & ours & \textbf{52.03} & \textbf{39.20}       \\ \hline
\multirow{2}{*}{EN/ZH} & xlm-roberta-base          & 31.36 & 20.89       \\
                       & ours & \textbf{60.13} & \textbf{46.80}       \\ \hline
\end{tabular}
}
\label{result}
\end{table}

\begin{figure}[h]
\centerline{\includegraphics[width=9cm]{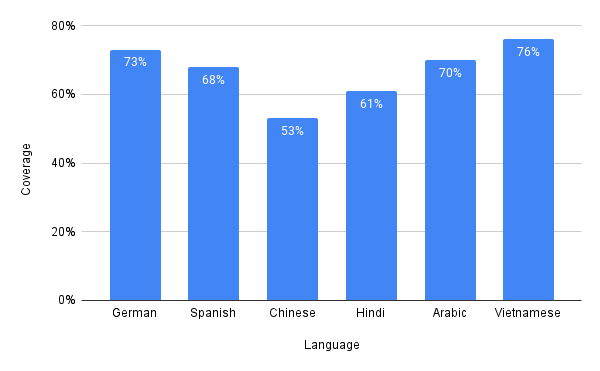}}
\caption{We first tokenize question text and knowledge triples in different languages. Then we calculate the coverage of tokens that appear in both questions and knowledge triples. That is to say, we calculate how much common tokens appear in tokenized triples. The results are visualized in this graph.
}
\label{cover}
\end{figure}

Tab.~\ref{result} presents the QA performance of the setting that context is English and question is changed from German to Chinese. Most obviously, we can observe that our method shows great improvement over the baseline method in terms of metrics used. However, the performance gain varies with the languages tested. To be more specific, among all 6 tested languages, we can see that improvement on German and Spanish is relatively low. Baseline model XLM-R-Base has the best performance when the questions are set to German and Spanish. Results on cross-lingual classification reported in~\cite{b3} indicate that the performance on Spanish and German are also the highest except English. We can also see from Fig.\ref{cover} that the token coverage of these two languages is only a little different from the others.
Take all these into consideration, and we think the limited improvement is caused by the deep understanding of German and Spanish in the baseline PLM. Significantly,  in the Arabic and Chinese cases, the baseline performance is the worst, indicating less knowledge storage and shallow understanding of the two languages. In the results on cross-lingual classification in~\cite{b3}, performance on Arabic and Chinese is obviously worse than in German and Spanish. After injecting knowledge, there are 18.48\% improvement on F1 score and 16.32\% improvement on Exact Match in Arabic and 28.77\% improvement on F1 score, 25.91\% improvement on Exact Match in Chinese. Dramatic performance gains prove the effectiveness of our method. When it comes to Hindi and Vietnamese, the same pattern is embodied.

There is 13.18\% improvement on F1 score and 12.00\% improvement on EM score on average. This indicates that using knowledge triples to enhance connections between different languages leads to significant performance gain in QA task. The improvement varies with the language tested since XLM-R-Base uses shared vocabulary.

\subsection{Case Studies}
Fig.~\ref{study} shows three example prediction cases of our model and the baseline model. In the first example, the question means 
\textit{``At this time, what finishing maneuver did he use?"}. The special noun mentioned in the question is apparently a low-frequency term. The model has first to realize that ``\begin{CJK}{UTF8}{gbsn}摔角终结技\end{CJK}'' means the same as ``finishing maneuver''. Then it needs to deal with the second low-frequency word ``Pedigree''. In the first example, the answer can be inferred from the sentence structure. That is to say, even without knowing the pedigree is a finishing maneuver, we can infer this is the answer. The baseline method predicts no answer, which is predictable since the term is rarely seen. Our method predicts almost the correct answer except definite article ``the''.
This case indicates that after injecting knowledge and aligning representations, the reasoning ability of PLM is highly enhanced. The second example is a medical-related problem. In this case, ``\begin{CJK}{UTF8}{gbsn}慢性支气管炎\end{CJK}'' means ``chronic cough''. However, the sentence containing the answer information does not include the entire term. It uses ``cough'' to represent ``chronic cough''. Again, the baseline method fails to find the correct answer, but our method manages to find a partial answer. This case proves that the improvement brought by our method is general, and our method can greatly enhance the ability to map different languages and reason. In the third example, a proper translation of the question is ``Professionals do not know whether there are different types of what?''. It's quite clear that this question and this context are easier than the previous two. Both our method and the baseline method find the correct answer, which indicates that the improvement brought by our method is stable.

\section{Future Work}
Multilingual wikidata triples are leveraged as a knowledge source in our work. By using link prediction technique, we can successfully inject external knowledge into PLMs and enhance the performance on G-XLT task. However, the results on cases where context is not English remain basically unchanged or even slightly drop. This phenomenon may be caused by the different storage of knowledge in each language. The challenge to better represent an entire context in low-resource languages remains unsolved. For future work, we will explore the following research directions: (1) In this work, we only study how to inject knowledge of two languages at once into PLMs. We will expand the capability of the proposed method via integrating more languages which is more practical.
With the increasing of languages used, new challenges might appear, like balancing the effects of different languages and how to bring new knowledge into PLMs effectively. (2) Each triple is viewed as an individual source of knowledge, and only knowledge bases are considered in our work. No relation between them or other knowledge sources are explored. In the future, We will also add more information in our joint model, such as relation paths in the knowledge base and auxiliary knowledge corpus. (3) We will try to use limited resources to solve the G-XLT problem better when the context is set to low-resource languages. (4) As listed in Fig.\ref{cover}, we can see the token coverage is at least over 50\%. If we can achieve almost the same performance gain with less data, out method can be more effective. Therefore, we will try to explore the relation between data coverage and the final performance gains.

\section{Conclusion}
% In this paper, we propose a novel knowledge injection and entity aligning method by exploiting the rich semantic information embedded in the textual representation of knowledge triples and enhancing the connection between different languages in PLMs. In this way, our model can bridge the language gap between different languages and promote the performance of G-XLT task. To capture different representation of an entity, we model it from two languages, and design automatic alignment process to align representations of the same entity. Extensive experiments on MLQA datasets show that our method significantly outperforms the state-of-the-art methods on the benchmark task of G-XLT. 

In this paper, we introduced a new G-XLT framework that fully exploited external knowledge bases. Our framework consists of two parts: (1) assembling triples and (2) knowledge injection. We train our model with triples assembled from two languages at once.
We show that it brings tremendous performance gains on English-context cases in MLQA. Further analysis shows that our method enriched the knowledge stored in each language and improved the reasoning ability of the baseline model. Our method consistently outperforms the baseline model on both simple and tough examples, which verifies the generality and stability of our model.
%(1) Entity linking may introduce noisy with incorrect entity annotations. We will explore techniques such as combining different entity linking methods to reduce these noises and improve the effectiveness of embedding.

\section*{Acknowledgments}
This work was supported in part by National Natural Science Foundation of China under Grant No. 61532010 and Beijing Academy of Artificial Intelligence (BAAI).

\vspace{12pt}

\end{document}